\documentclass[10pt,twocolumn,letterpaper]{article}

\usepackage{cvpr}              
\usepackage{makecell}
\usepackage{multirow}
\usepackage{graphicx}

\usepackage{lineno}

\definecolor{cvprblue}{rgb}{0.21,0.49,0.74}
\usepackage[pagebackref,breaklinks,colorlinks,allcolors=cvprblue]{hyperref}

\def\paperID{14830} 
\def\confName{CVPR}
\def\confYear{2025}

\title{Towards Smart Point-and-Shoot Photography}

\author{
Jiawan Li$^{1,2,3,4}$ , Fei Zhou$^{1,2,3,4,*}$ , Zhipeng Zhong$^{5}$ , Jiongzhi Lin$^{1,2,3,4}$ , Guoping Qiu$^{6,7}$\thanks{Corresponding author}\\
$^1$ Shenzhen University,$^2$ Guangdong Provincial Key Laboratory of Intelligent Information Processing \\
$^3$Guangdong-Hong Kong Joint Laboratory for Big Data Imaging and Communication\\
$^4$ Shenzhen Key Laboratory of Digital Creative Technology\\
$^5$ Loughborough University, $^6$ University of Nottingham,$^7$ Everimaging Ltd\\
{\tt\small lijiawan2022@email.szu.edu.cn,flying.zhou@163.com,2019281030@email.szu.edu.cn}\\
{\tt\small Z.Zhong@lboro.ac.uk, guoping.qiu@nottingham.edu.cn }
}

\begin{document}
\maketitle
\begin{abstract}
\label{sec:abstract}
Hundreds of millions of people routinely take photos using their smartphones as point and shoot (PAS) cameras, yet very few would have the photography skills to compose a good shot of a scene. While traditional PAS cameras have built-in
functions to ensure a photo is well focused and has the right brightness, they cannot tell the users how to compose the best shot of a scene. In this paper, we present a first of its kind smart point and shoot (SPAS) system to help users to take good photos. Our SPAS proposes to help users to compose a good shot of a scene by automatically guiding the users to adjust the camera pose live on the scene. We first constructed a large dataset containing $320K$ images with camera pose information from 4000 scenes. We then developed an innovative CLIP-based Composition Quality Assessment (CCQA) model to assign pseudo labels to these images. The CCQA introduces a unique learnable text embedding technique to learn continuous word embeddings capable of discerning subtle visual quality differences in the range covered by five levels of quality description words $\{bad, poor, fair, good, perfect\}$. And finally we have developed a camera pose adjustment model (CPAM) which first determines if the current view can be further improved and if so it outputs the adjust suggestion in the form of two camera pose adjustment angles. The two tasks of CPAM make decisions in a sequential manner and each involves different sets of training samples, we have developed a mixture-of-experts model with a gated loss function to train the CPAM in an end-to-end manner. We will present extensive results to demonstrate the performances of our SPAS system using publicly available image composition datasets.
\end{abstract}    
\section{Introduction}
\label{sec:intro}

Traditional Point-and-Shoot (PAS) cameras have built-in functions such as autofocus, autoexposure, and auto-flash to ensure a photograph is well focused and has the right brightness. However, these PAS cameras cannot tell the users how to compose the best shot of a scene. It is estimated that there are over 7 billion smartphones worldwide and every one is a PAS camera (in the context of this paper, smartphone and camera are used interchangeably). Although almost every smartphone user would routinely use their phones to take photos, very few would have the photography skill to compose a good shot of a scene. In this paper, we present a solution that automatically guides smartphone users to compose the best shot live on a scene. 
 \begin{figure}[t]
  \centering
   \includegraphics[width=0.98\linewidth]{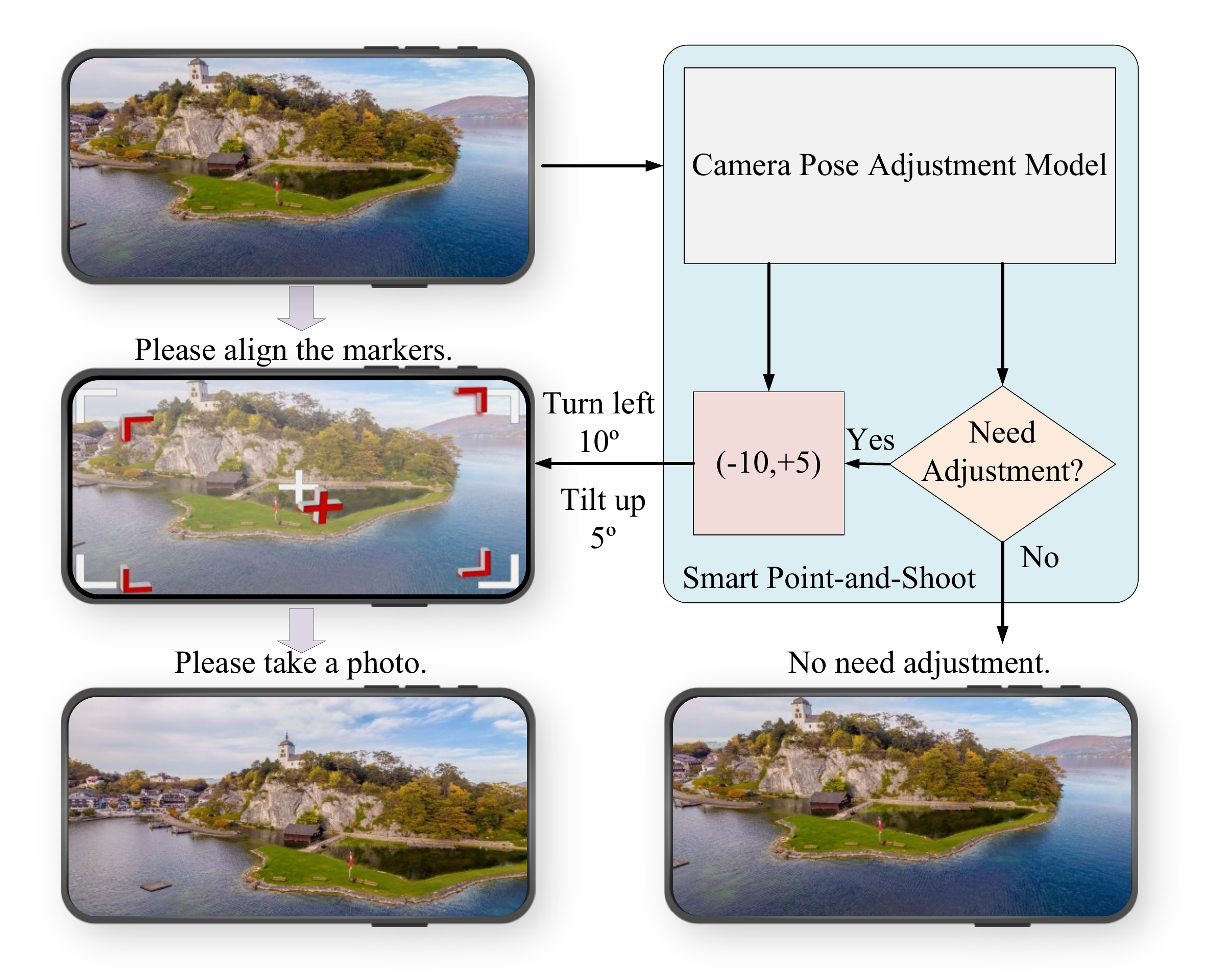}

   \caption{Given a view composed by the user, our Smart Point-and-Shoot (SPAS) system can predict camera pose adjustment suggestions so that the photo captured after applying the adjustment will have a better composition.}
   \label{fig:1}
\end{figure}

For a given scene of interest and starting from an initial view a user points to, our Smart Point-and-Shoot (SPAS) system automatically recommends camera pose adjustment strategies and guide the user to rotate the camera upwards, downwards, rightwards or leftwards until the camera points to the best shot. In contrast to existing literature on automatic picture composition which is a post-processing procedure of cropping a photo that has already been taken from a fixed view, our SPAS is the first system that enables users to compose the best shot of a scene by guiding the users to adjust the camera pose live on the scene.

As shown in \Cref{fig:1}, given an initial view, the camera pose adjustment model (CPAM) first evaluates whether the composition can be improved. If so, it predicts how the camera pose should be adjusted. Specifically, let $\theta$, $\varphi$, and $\gamma$ respectively denote the yaw, pitch, and roll angles of a camera pose $P = (\theta, \varphi,\gamma)$. Because it is unusual to roll the camera during shooting, it is reasonable to assume that the roll angle is fixed. The CPAM therefore suggests how to rotate along the vertical axis (change yaw angle $\theta$)  and how to rotate along the horizontal axis (change pitch angle $\varphi$).  By providing camera pose adjustment suggestions during the shooting process, we can help the users to effectively improve the composition and take a good shot of the scene. The challenge now is how to construct the camera pose adjustment model (CPAM). 

\begin{figure}[t]
  \centering
  \includegraphics[width=0.98\linewidth]{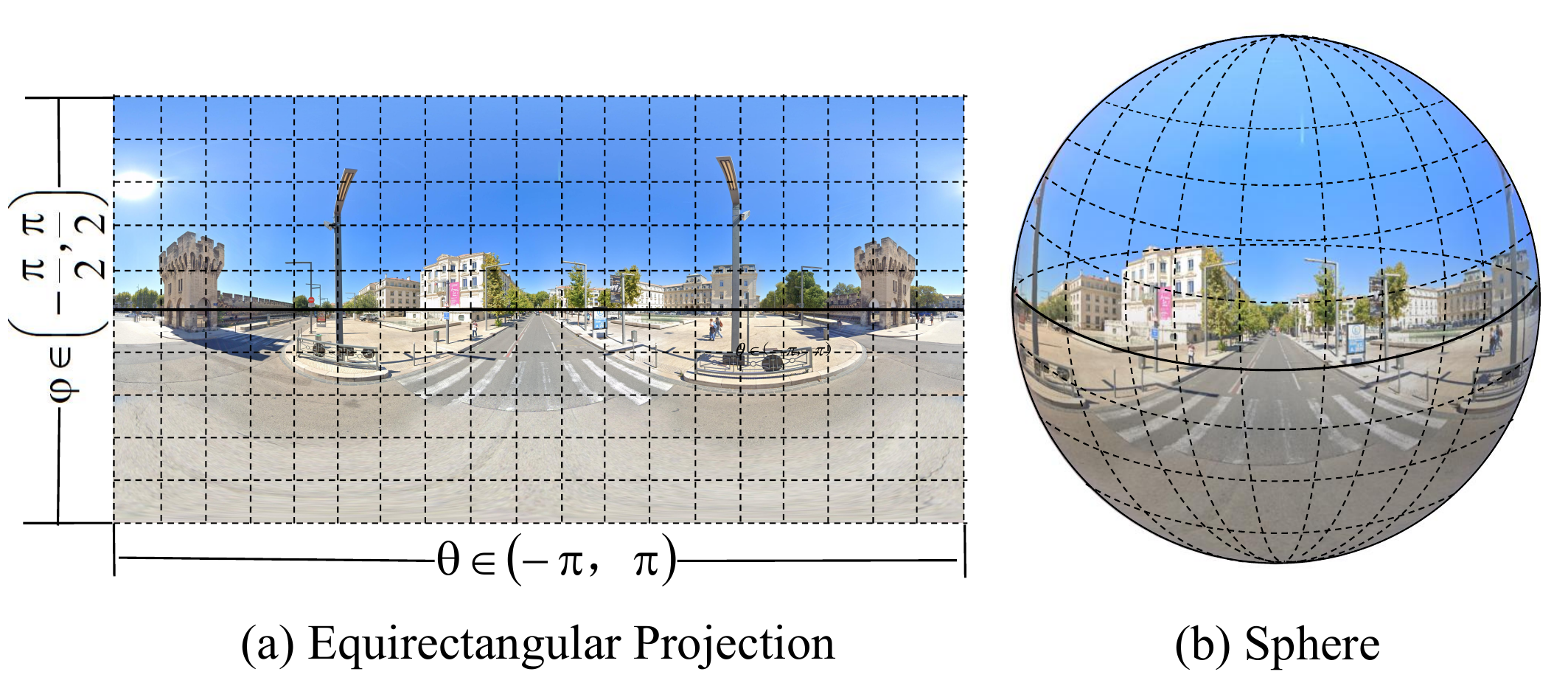}

   \caption{Geometric explanation of the relationship between ERP (a) and the sphere (b).}
   \label{fig:2}
\end{figure}

First of all, we require a suitably annotated dataset and then use the data to construct an intelligent model that can first determine if a given view's composition can be improved and if so how the camera pose should be changed in order to obtain an improved shot. The challenge of obtaining a large enough dataset is significant. Manually acquiring images of different camera poses from a variety of scenes and then annotate them with composition scores will be extremely time consuming and therefore is impracticable. In terms of the CPAM itself, it needs to perform sequential decision making on two tasks. It is therefore particularly important to model the relationship between the tasks  to avoid conflicts arising from task discrepancies. In this paper, we have developed practical solutions to these problems. 

To construct a dataset for the problem, we first take advantage of the availability of $360^{\circ}$ images of Google Street View\footnote{\url{https://www.google.com/streetview/}}. By exploiting the geometric explanation of the relationship between the equirectangular projection (ERP) of the $360^{\circ}$ image and the sphere (see \Cref{fig:2}), we discover that a panoramic image in the ERP format can be mapped onto the surface of a unit sphere. This mapping creates a complete $360^{\circ}$ photographic environment where spherical coordinates (longitude $\theta$ and latitude $\varphi$) naturally correspond to the orientation of a virtual camera positioned at the sphere's center. Through this geometric correspondence, we can precisely control the camera's viewing direction using these spherical coordinates, enabling the generation of sample views with well-defined camera poses $(\theta,\varphi)$ via perspective projection. Based on this observation, we create the Panorama-based Composition Adjustment Recommendation dataset (PCARD). As shown in \Cref{tab:dataset}, the new PCARD contains $320K$ images with camera pose information from 4000 scenes. As far as we know, this is the first dataset created from the $360^{\circ}$ images of Google Street View where each image contains the camera pose information. We will use the PCARD to develop a smart point and shoot (SPAS) solution.

\begin{table}[t]  
    \centering  
    \resizebox{1\columnwidth}{!}{  
    \begin{tabular}{cccccc}  
    \toprule  
    Dataset & Year & Label & Scenes & \makecell{Candidate\\ Views}  & \makecell{Camera \\Pose}  \\
    \hline  
    ICDB\cite{ICDB} & 2013 & Best & 950 & 1 & N/A \\
    HCDB\cite{HCDB} & 2014 & Best & 500 & 1 & N/A \\
    GNMC\cite{GNMC} & 2022 & Best & 10000 & 5 & N/A \\
    SACD\cite{SACD} & 2023 & Best & 2777 & 8  & N/A  \\
    \hline  
    FCDB\cite{r1} & 2017 & Rank & 1536 & 18 & N/A \\
    CPC\cite{r4} & 2018 & Rank & 10800 & 24 & N/A \\
    GAICv1\cite{r7} & 2019 & Score & 1236 & 86 & N/A \\
    GAICv2\cite{r5} & 2020 & Score & 3336 & 86 & N/A  \\
    UGCrop5K\cite{r10} & 2024 & Score & 5000 & 90 & N/A  \\
    PCARD(Ours) & 2024 & Score & 4000 & 81 & 324000 \\
    \bottomrule  
    \end{tabular}}  
    \caption{Image Composition datasets and PCARD.}  
    \label{tab:dataset}  
\end{table}

For the $320K$ images in PCARD, it is necessary to assign each a quality label. Again manual approach is impractical. Instead we resort to images with composition quality ratings such as those in \cite{r5} to train a labeler to assign pseudo composition score labels to these images. One of the major challenges in developing the pseudo labeler is that neigbouring views have large overlapping regions and are very similar, therefore the labeler needs to have the ability to distinguish images with subtle differences. In this paper, we take full advantage of large language models (LLM) and have developed a CLIP-based Composition Quality Assessment  (CCQA) model.
As CLIP is sensitive to the choice of prompts and text descriptions of nuance visual differences are difficult, we abandon traditional subjective prompt settings in favor of learnable text prompts. We have developed an effective method that learns continuous word embeddings capable of discerning subtle visual quality differences in the range covered by five levels of quality description words $\{bad, poor, fair, good, perfect\}$. 

The Camera Pose Adjustment model (CPAM) performs two tasks in a sequential manner. Logically, it needs to first determine if the current view can be further improved and if so it then outputs the adjust suggestion in the form of two pose adjustment angles. This is a multitask learning problem but logically the decisions must be made in a sequential manner. Also, unlike normal multitask learning, the two learning tasks involve different training samples with one involves the full set and the other a subset of the training samples. To tackle this problem, we have developed a mixture-of-experts model with a gated loss function to train the CPAM in an end-to-end manner. 
In summary, this paper makes 4 major contributions:
\begin{itemize}
    \item We present a first of its kind smart point and shoot (SPAS) system to help the billions of smartphone users to take good photographs. Our SPAS is the first in the literature that proposes to help users to compose a good shot of a scene by automatically guiding the users to adjust the camera pose live on the scene. 
    \item 
    We have constructed a large dataset containing $320K$ images with camera pose information from 4000 scenes by exploiting the availability of $360^\circ$ images of Google Street View. This dataset which will be made publicly available and can  be used for the task in this paper as well as other applications.  
    \item 
    We have developed an innovative CLIP-based Composition Quality Assessment (CCQA) model. The CCQA introduces a unique learnable text embedding technique to learn continuous word embeddings capable of discerning subtle visual quality differences in the range covered by five levels of quality description words $\{bad, poor, fair, good, perfect\}$.
    \item
    We have developed a camera pose adjustment model (CPAM) which first determines if the current view can be improved and if so it outputs the adjust suggestion in the form of two camera pose adjustment angles. The two tasks of CPAM make decisions in a sequential manner and each involves different sets of training samples, we have developed a mixture-of-experts model with a gated loss function to train the CPAM in an end-to-end manner.

\end{itemize}

\section{Related Work}
\label{sec:related_work}
\textbf{Image Composition dataset.} 
For photo recommendation tasks, there exist some image cropping datasets \cite{ICDB, HCDB, r1, GNMC, r4, r7, r5, SACD, r10} that can be categorized into two groups based on their annotation styles, as shown in \Cref{tab:dataset}.
More details can be seen in the Supplementary Material.

\textbf{Aesthetic-guided image composition.} Image aesthetic quality assessment aims to quantify image aesthetic values, while image composition focuses on finding the most aesthetic view. While prior works \cite{ICQA_1, ICQA_2, ICQA_3, ICQA_4, ICQA_5} learn aesthetic-related features to evaluate composition quality, they lack recommendation capabilities. Instead, image cropping, which aims to find the most aesthetic sub-region through cropping boxes, has emerged as a promising direction. Existing methods \cite{r1,r2,r3,r4,r5,r6,r7,r8,r9,r10,r11,r12,r13,r14,r15,r16,r17,r18} generally fall into two categories: score-based methods \cite{r1,r2,r3,r4,r5,r6,r7,r8,r9,r10,clipcrop} that evaluate candidate views using learned aesthetic knowledge, and coordinate regression-based \cite{r11,r12,r13,r14,r15,r16,r17,r18} methods that directly predict optimal cropping boxes through various learning strategies.

Although previous methods have achieved good results for cropping-based image composition tasks, image cropping is a post-processing exercise applied to already captured images where the viewpoints have already been fixed. It is not applicable in scenarios where the photographer needs to adjust the camera pose or position to capture the best view of a scene. In this work, we present a framework that automatically provides photographers with camera pose adjustment directions and guides the photographers to take the best shot of a given scene.

\section{Problem Definition and Overview}
\label{sec:problem_definition}
\begin{figure}[t]
    \centering
    \includegraphics[width=0.99\linewidth]{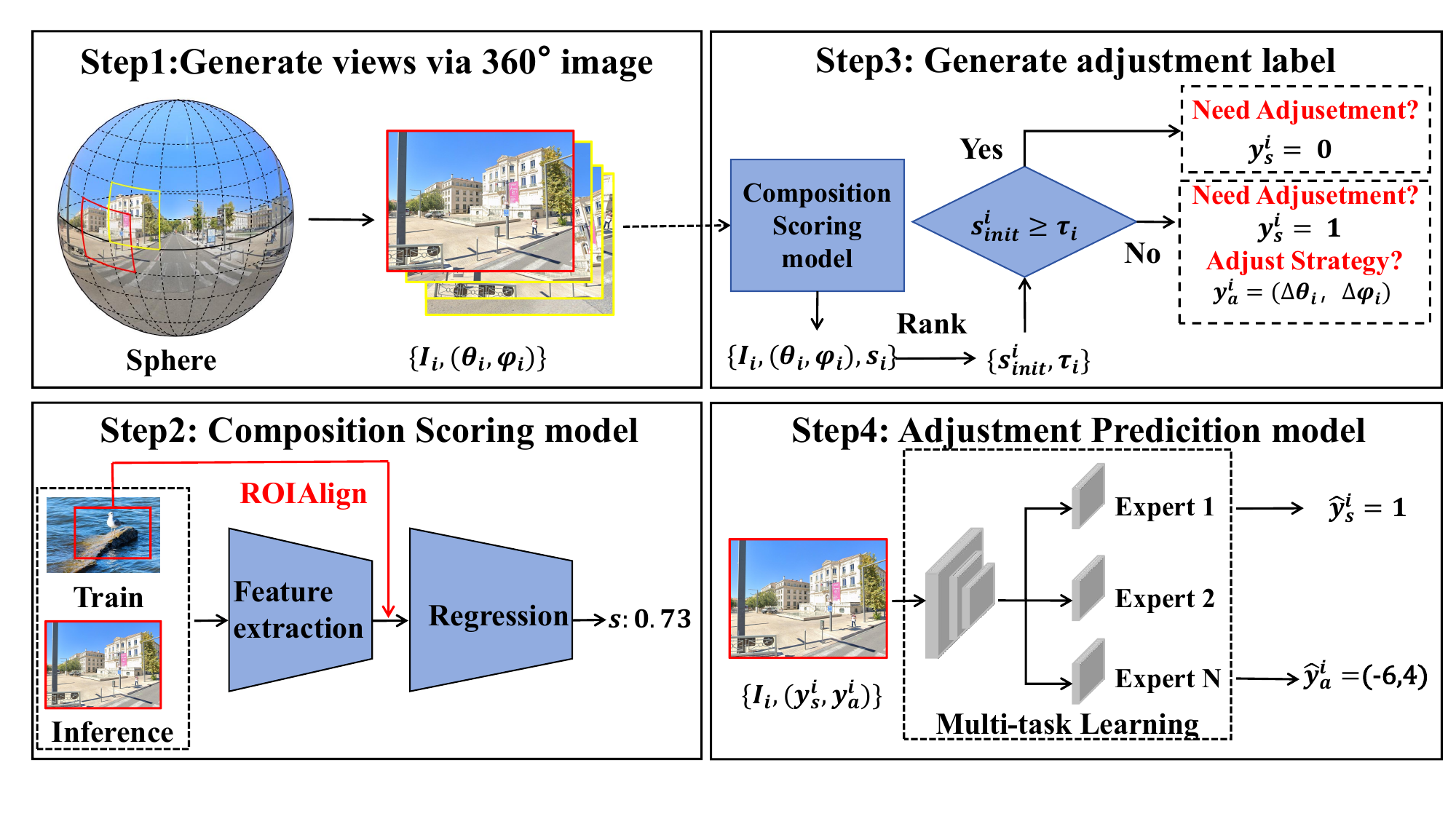}
    \caption{The overview of our method. Using perspective projection, we generate views from 360° images with camera poses (Step 1). We train a composition scoring model to evaluate image composition quality (Step 2) and design a composition quality score-guided method to generate camera pose adjustment labels (Step 3). Finally, a sequential multi-task MoE network predicts camera adjustments to improve image composition (Step 4).}
    \label{fig:3}
\end{figure}
In general, a photographer assesses an initial view through the viewfinder and then adjusts the camera pose utilizing 3 degrees of freedom in the 3D world space (yaw $\theta$, pitch $\varphi$, roll $\gamma$) to take the best shot. 

Given an initial view $\boldsymbol{I}_{{init}}^{i}$ of the $i^{th}$ scene, and a camera pose adjustment prediction model $f(\cdot)$, the problem can be formulated by
\begin{equation}
\left(\widehat{\boldsymbol{y}}_{{s}}^{i},\widehat{\boldsymbol{y}}_{{a}}^{i}\right)=f\left(\boldsymbol{I}_{{init}}^{i}\right)
\end{equation}
where $\widehat{\boldsymbol{y}}_{{s}}^{i}$ and $\widehat{\boldsymbol{y}}_{{a}}^{i}$ respectively represent the suggestion output and the adjustment output. $\widehat{\boldsymbol{y}}_{{s}}^{i}$ indicates whether the composition of an initial view $\boldsymbol{I}_{{init}}^{i}$ can be improved. If the composition can be improved, the adjustment predictor predicts the suitable camera pose adjustment strategy, which is $(\Delta \theta_i, \Delta\varphi_i,\Delta\gamma_i)$. In practice, it is unusual to roll the camera during the photography process, we therefore fix the roll angle $\gamma$. The camera pose and the camera pose adjustment strategy can be further simplified to ($\theta_i$, $\varphi_i$) and ($\Delta \theta_i, \Delta\varphi_i$). $\theta_i \in[-180^{\circ}, 180^{\circ}]$ and $\varphi_i \in[-90^{\circ}, 90^{\circ}]$.  $\Delta \theta \in[-180^{\circ},180^{\circ}]$ represents the camera pose rotating left or right around the vertical axis, with rightward rotation being positive. $\Delta \varphi \in[-180^{\circ},180^{\circ}]$ represents the camera pose rotating up or down around the horizontal axis, with upward being positive.
The pipeline of the whole approach is illustrated in \Cref{fig:3}. First, to train the camera pose adjustment prediction model $f(\cdot)$, we create the Panorama-based Composition Adjustment Recommendation dataset $\mathcal{D_{PCARD}}=\left\{\boldsymbol{I}_{{init}}^{i},\boldsymbol{y}_{{s}}^{i},\boldsymbol{y}_{{a}}^{i}\right\}_{i=1}^{N_{\text {scene}}}$ and present a pseudo-labeling method guided by composition quality scores to generate the camera pose adjustment labels $(\boldsymbol{y}_{{s}}^{i},\boldsymbol{y}_{{a}}^{i})$ (\cref{sec:dataset}). Specially, we propose a CLIP-based Composition  Quality Assessment (CCQA) model $h(\cdot)$ to evaluate the composition quality of views $\boldsymbol{I}$ (\cref{sec:analysis_ccqa}). Subsequently, the Camera Pose Adjustment model (CPAM) $f(\cdot)$ is illustrated in \cref{sec:adjustment_model}.

\section{PCARD Database}
\label{sec:dataset}

\begin{figure}[t]
    \centering
    \includegraphics[width=0.99\linewidth]{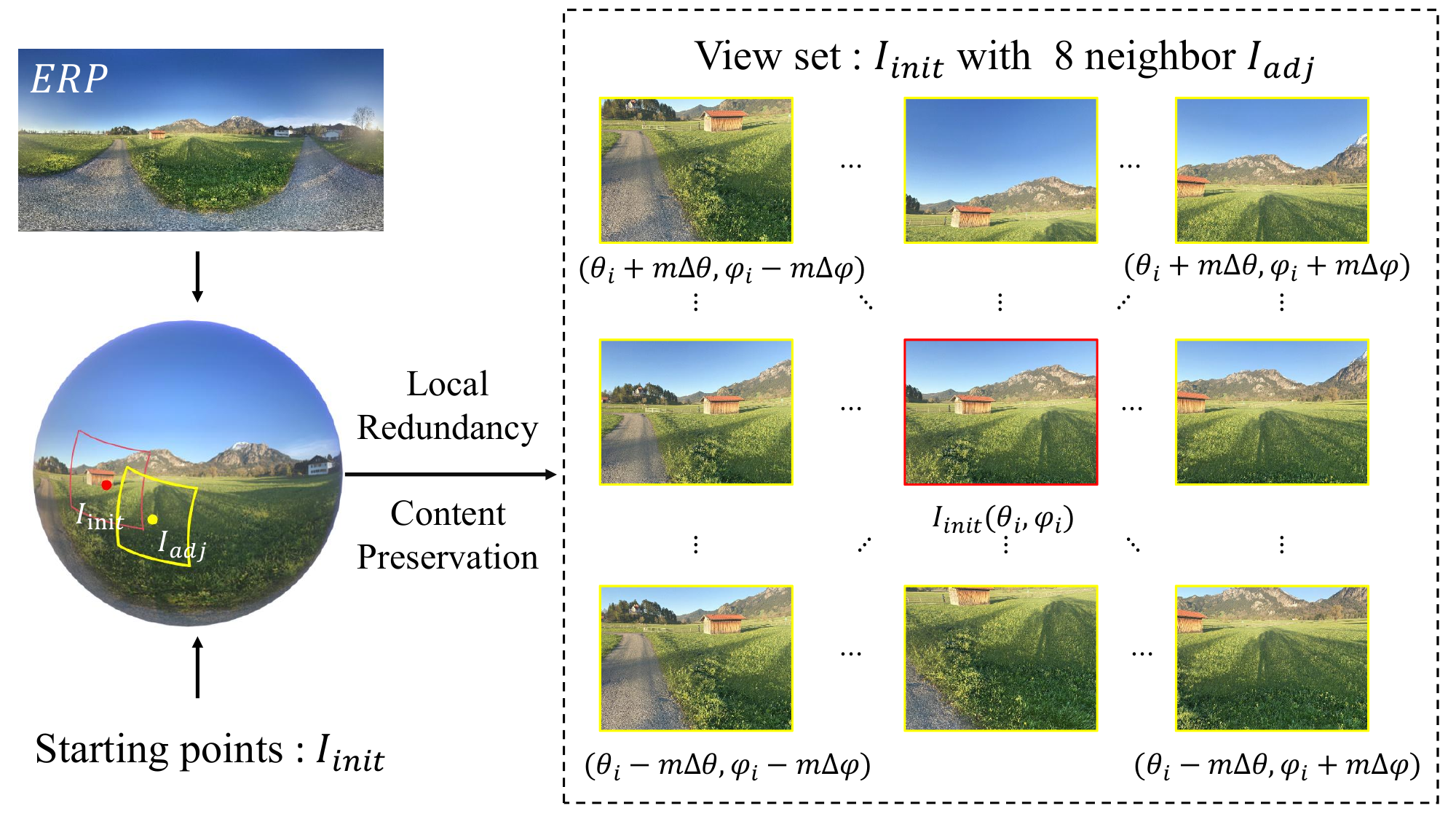}
    \caption{A multi-angle view generation method based on the $360^{\circ}$ images.}
    \label{fig:4_genrate_view}
\end{figure}
\subsection{Formulation}
\label{sec:fromulation}
As shown in \Cref{fig:4_genrate_view},
given an ERP image with spatial resolution $H \times W$, we transform it into a unit sphere $S^3$ with $S^2$ as its surface. Every point $(\theta, \phi) \in S^2$ is uniquely defined by its longitude $\theta \in[-\pi, \pi]$ and latitude $\phi \in[-\pi/2,\pi/2] $. In the spherical domain, this can be expressed as:
\begin{equation}
\left\{\begin{array}{l}
\theta=\frac{2 \pi u}{W}-\pi \\
\varphi=\frac{-\pi v}{H}+\frac{\pi}{2}
\end{array}\right.
\end{equation}
where $u \in[1,W]$ and $v \in[1,H]$.
We assume that a virtual pinhole camera is positioned at the center of the sphere $S^3$. The visual content is then captured as a planar view $\boldsymbol{I}_{{init}}^{i}$ that is determined by the viewing angle $(\theta_{{init}}^{i},\varphi_{{init}}^{i})$, and field of view $(fov_x ,fov_y)$ of the camera through perspective projection\cite{hajjami2020arucomni}. By adjusting the camera pose we generate candidate views 
$\boldsymbol{{I}}_{adj}^{i}=\{I_i^{j},(\theta_{i}^{j},\varphi_{i}^{j}) \}_{j=1}^{M}$, where $M=\frac{360}{\Delta \theta}\times \frac{180}{\Delta\varphi}$ which is the number of candidate views, $\Delta \theta$ and $\Delta\varphi$ are the view adjustment step-sizes. Then the search space $M$ is efficiently reduced by exploiting the \textit{Content Preservation} and \textit{Local Redundancy} properties. 
And to complete the dataset construction process, we propose a pseudo-labeling method guided by composition quality scores to generate the camera pose adjustment labels $(\boldsymbol{y}_{{s}}^{i},\boldsymbol{y}_{{a}}^{i})$ for the candidate views.

\textbf{Content Preservation}. Generally speaking, the adjusted view $\boldsymbol{I}_{{adj}}^{i}$ should preserve the main content of the initial view $\boldsymbol{I}_{{init}}^{i}$ to maintain the photographer's intended subject.
Hence, we constrain the overlapping area between the adjusted next view $\boldsymbol{I}_{{adj}}^{i}$ and the initial view $\boldsymbol{I}_{{init}}^{i}$ to be no smaller than a certain proportion of $\boldsymbol{I}_{{init}}^{i}$. Note that it is directly defined on a sphere (the $360^{\circ}$ images) rather than ERP or the tangent plane \cite{zhao2020spherical}.
\begin{equation}
\text{ SphOverlap }\left(S_{\text {adj }}, S_{\text {init }}\right)=\frac{A(S_{\text {adj }} \cap S_{\text {init }})}{A(S_{\text {init }})}>\lambda
\end{equation}
where ${S}_{adj}$ and ${S}_{init}$ represent the spherical rectangles corresponding to $\boldsymbol{I}_{{adj}}^{i}$ and $\boldsymbol{I}_{{init}}^{i}$ in the $360^{\circ}$ images respectively, $A(\cdot)$ is the area of the shape and $\lambda \in\left[0.5, 1\right)$.

\textbf{Local Redundancy}. Adjusting the camera pose to improve image composition is a problem with local redundancy because suboptimal solutions are also acceptable. Based on Moore neighborhood theory\cite{Moore}, we design a sampling matrix that captures 8 neighboring views around the current camera pose $(\theta_i,\varphi_i)$ at varying distances controlled by a multiplier $m$, as shown in \Cref{fig:4_genrate_view}. To efficiently remove redundant candidate views, we set the sampling step sizes $\Delta\theta$ = $\Delta\varphi = 5^{\circ}$ following \cite{360view_5}.

More detailed mathematical calculations can be found in the Supplementary Material.
\subsection{Database Construction}
\label{sec:4_database_construction}

We selected 20 countries from Street View Download 360 \footnote{\url{https://svd360.com/}}, 
with an average of 8 cities per country, resulting in a total download of over 150K $360^{\circ}$ images in equirectangular projection (ERP) format. We designed a web player based on a Web3D library Three.js\footnote{\url{https://threejs.org/}} to achieve $360^{\circ}$ playback of the panoramic images. This allows us to inspect the panoramic images for obvious distortion or damage, and if none are present, select suitable initial views and record their camera poses.
In the end, we retained 4,000 high-quality panoramic images.
For each panoramic image, we first generate the initial view $\boldsymbol{I}_{{init}}^{i}$ according to the pre-recorded camera poses $(\theta_{{init}}^{i},\varphi_{{init}}^{i})$ and then generate candidate views $\boldsymbol{{I}}_{adj}^{i}=\{I_i^{j},(\theta_{i}^{j},\varphi_{i}^{j}) \}_{j=1}^{M}$ following the Content Preservation and Local Redundancy in \cref{sec:fromulation}, where $M$ is the size of candidate view set. In our final dataset, on average, $M = 81$ which we believe is a reasonable size for learning image composition. 

\subsection{Label Generation}
\label{sec:4_pseudo_label_generation}
We propose a labeling method guided by aesthetic scores to generate the camera pose adjustment labels $(\boldsymbol{y}_{{s}}^{i},\boldsymbol{y}_{{a}}^{i})$ of the view $\boldsymbol{I}_{{init}}^{i}$. To do this, we have designed a CLIP-based Composition Quality Assessment (CCQA) model which will be described in \cref{subsec:CCQA}. Given an initial view $\boldsymbol{I}_{{init}}^{i}$ with camera pose $(\theta_{{init}}^{i},\varphi_{{init}}^{i})$ and its corresponding candidate views $\boldsymbol{{I}}_{adj}^{i}=\{I_i^{j},(\theta_{i}^{j},\varphi_{i}^{j}) \}_{j=1}^{M}$, we use the CCQA model $h(\cdot)$ to assign numerical composition quality ratings to views: $s = h(I)$. We denoted that $s_{{init}}^{i}$ represents the score for the initial view $\boldsymbol{I}_{{init}}^{i}$ while $s_{{adj}}^{i}$ contains the scores for the candidate views $\boldsymbol{{I}}_{adj}^{i}$.

Then, we calculate the adaptive threshold $\tau_i$ for each scene $i$. This threshold is determined by ranking the scores of the candidate views in descending order and selecting the $N^{th}$ score: $\tau_{i} = Top_{N}({s_{{adj }}^{i}})$
where $N$ is a fixed percentage of the total number of candidates $M$. In practice, we set $N= 25\%$, the detailed information will be discussed in Supplementary.

Finally, we generate the suggestion label $\boldsymbol{y}_{{s}}^{i}$ and adjustment label $\boldsymbol{y}_{{a}}^{i}$ leveraging the adaptive threshold $\tau_i$:
\begin{equation}
{\boldsymbol{y}}_{{s}}^{i}=\left\{\begin{array}{l}
1, \text { if } s_{{init}}^{i}<\tau_{i} \\
0, \text { if } s_{{init}}^{i}>=\tau_{i}
\end{array}\right.
\label{eq:suggestion_label}
\end{equation}
\begin{equation}
\boldsymbol{y}_{{a}}^{i}=\left\{\begin{array}{cc}
\left(\theta_{\text {best }}^i, \varphi_{\text {best }}^i\right)-\left(\theta_{\text {init }}^i, \varphi_{\text {init }}^i\right), & \text { if } \boldsymbol{y}_{{s}}^{i}=1 \\
(0,0), & \text { otherwise }
\end{array}\right.
\label{eq:adjustment_label}
\end{equation}
where $(\theta_{{best}}^{i},\varphi_{{best}}^{i})$ represents the camera pose of the candidate view with the highest composition quality score.

\section{CLIP-based Composition Quality Assessment}
\label{sec:analysis_ccqa}

\label{subsec:CCQA}
\begin{figure}[t]
    \centering
    \includegraphics[width=1\linewidth]{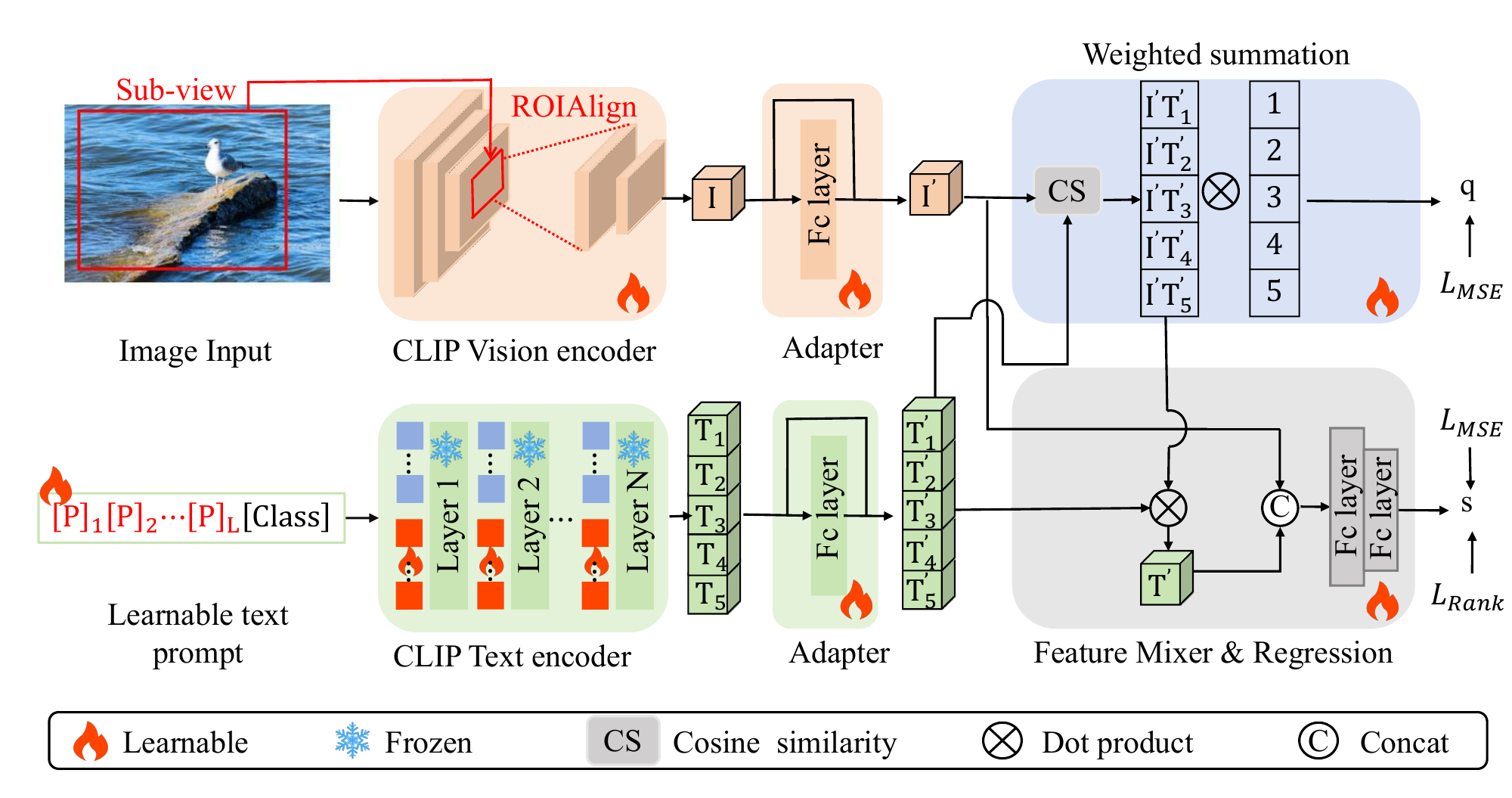}
    \caption{CLIP-based Composition Quality Assessment model.}
    \label{fig:CCQA}
\end{figure}
We introduce our CLIP-based Composition Quality Assessment (CCQA) model illustrated in \Cref{fig:CCQA}. The model is trained on the GAICv2 dataset \cite{r5} that pairs each image $x$ with multiple cropping view $v$ and their corresponding composition quality scores $s$. 

\textbf{Image encoder}. Given an image $x$ and a set of view $v$, the image encoder creates global feature maps from $x$ by a trainable CLIP image encoder's first 3 blocks, then uses RoIAlign to extract sub-view feature which are further encoded by CLIP's final block to obtain sub-view embeddings that denoted as visual embedding $I$.

\textbf{Learnable prompt}. The design of prompts can greatly impact performances. CLIP is sensitive to the choice of prompts, therefore, we abandon traditional subjective prompt settings in favor of a learnable prompt strategy. These learnable text prompts $T$ are defined as follows:
\begin{equation}
T=[P]_1[P]_2 \ldots[P]_L[Class]
\end{equation}
Each $[P]_l(l \in\{1, \ldots, L\})$ is a learnable word embedding in the text prompt templates with the same 512 dimensionality as the CLIP word embedding, $L$ represents the number of context tokens. $Class$ is one of five-level quality description words $\{bad, poor, fair, good, perfect\}$. 

\textbf{Feature adapters and Weighted summation}. We introduce learnable feature adapters to better leverage CLIP's prior knowledge and enhance visual-text feature synergy. The adapted features $I^{\prime}$ and $T^{\prime}$ are obtained by applying residual adaptation and normalization to the visual embedding $I$ and text embeddings $T$  respectively.

The quality weights $W_i$ are computed by applying softmax to the cosine similarities between adapted image feature $I^{\prime}$ and five class prompts $\{{T_i}^{\prime }\}_{i=1}^5$\cite{cvpr2023,icml2024}.
\begin{equation}
W_i=\frac{\exp \left(I^{\prime \top} {T_i}^{\prime} / \sigma \right)}{\sum_{j=1}^5 \exp \left(I^{\prime \top} {T_j}^{\prime} / \sigma\right)},
\end{equation}
where $\sigma$ is the temperature parameter. The assessment score $q$ of the given image is calculated as: 
\begin{equation}
q=\sum_{i=1}^5 W_i \times C_i,
\label{eq:weighted_sum}
\end{equation}
where $\{{C_i}\}_{i=1}^5$ are the numerical scores of the five-level quality description words which are set to ${1, 2, 3, 4}$ and $5$ with a lower numerical value corresponds to a lower quality class word.  

\textbf{Feature mixers and regression}.  To better enable the CLIP features to discern subtle differences in aesthetic quality across a series of similar photos, we obtain the weighted text features $F_t$ through calculating the dot product between the quality weights $W_i$ and the adapted text features $\{{T_i}^{\prime }\}_{i=1}^5$ of the five prompts:
\begin{equation}
F_t=\sum_{i=1}^5 W_i \times {T_i}^{\prime},
\end{equation}
The final score $\hat{s}$ is predicted by passing the concatenated weighted text features $F_t$ and adapted image features $I^{\prime}$ through an MLP.

\textbf{Optimization}. 
The CCQA uses a multi-task loss function. The first task enforces the predicted scores to be close to their ground truth scores: 
\begin{equation}
\mathcal{L}_{1} =\frac{1}{N} \sum_{i=1}^N(\hat{s_i}-s_i)^2
\end{equation}
where $\hat{s_i}$ and $s_i$ represent the score predicted by CCQA and the ground truth.

The second task enforces the predicted scores of different views to have the same ranking order as that of the ground truth scores.
We therefore also incorporate a ranking loss $\mathcal{L}_{2}$ to explicitly model the ranking relationship.
\begin{equation}
\mathcal{L}_{2}=\frac{\sum_{i, j} \max \left(0,-sign\left({s_i}-s_j\right)\left(\left(\hat{s_i}-\hat{s_j}\right)-\left(s_i-s_j\right)\right)\right).}{N(N-1) / 2}
\end{equation}
where $sign(\cdot)$ is the standard sign function.

The third task $\mathcal{L}_{3}$ enforces consistency between $q$ (cosine similarity-based weighting computed according to \cref{eq:weighted_sum}), and the ground truth scores $s_i$. 
\begin{equation}
\mathcal{L}_{3}=\frac{1}{N} \sum_{i=1}^N(q_i-s_i)^2
\end{equation}

The total loss function can be summarized as
\begin{equation}
\mathcal{L}_{CCQA}= \mathcal{L}_{1} + \mathcal{L}_{2} + \alpha * \mathcal{L}_{3}
\end{equation}
where the hyperparameters $\alpha$ is used to balance different losses (set to $0.1$ in this paper).


\section{Camera Pose Adjustment Model}
\label{sec:adjustment_model}

\begin{figure}[t]
    \centering
    \includegraphics[width=1\linewidth]{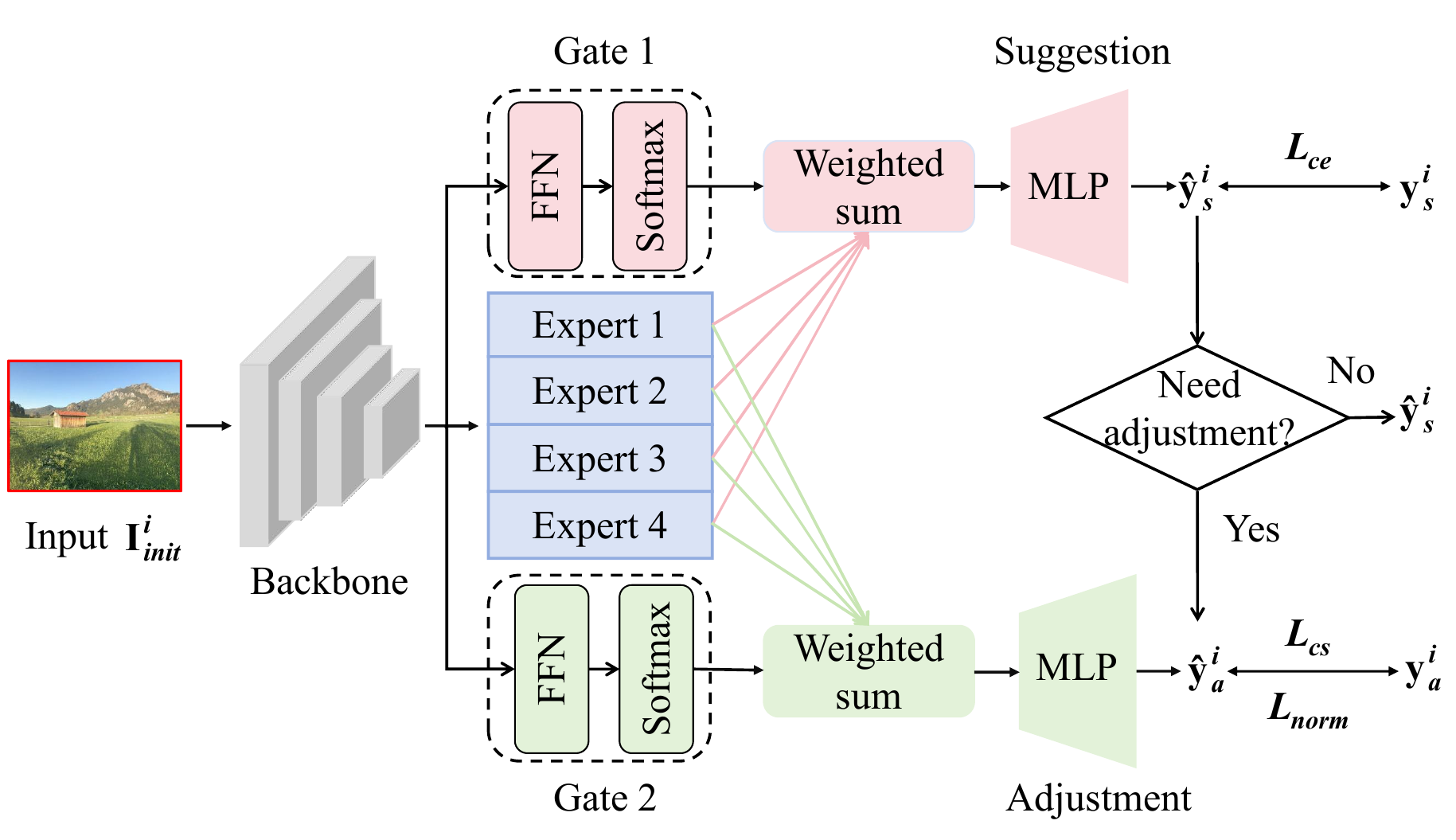}
    \caption{Camera Pose Adjustment model.}
    \label{fig：adjustment_model}
\end{figure}

Given an image, the Camera Pose Adjustment model (CPAM) $f(\cdot) $ produces two outputs. Firstly, the output of the suggestion predictor $\widehat{\boldsymbol{y}}_s^i$ predicts whether a view adjustment should be performed. Suggestion predictor is a binary classification head. 
Secondly, the output of the adjustment predictor $\widehat{\boldsymbol{y}}_a^i$ predicts how to adjust the camera pose when the suggestion predictor indicates an adjustment is needed. The adjustment predictor has a regression head respectively predicting the variables $\Delta\theta$ and $\Delta\varphi$.

A key challenge is the sequential dependency between these tasks - the adjustment prediction is only meaningful when the suggestion predictor indicates adjustment is needed. This creates an imbalanced training scenario where only a subset of samples contribute to the adjustment task, potentially causing conflicts between tasks due to different sample spaces and gradient frequencies.

To resolve this problem, we adopt a multi-gate mixture of experts architecture, which allows each task to adaptively control parameter sharing through task-specific gates, enabling the model to learn task-specific features while maintaining shared knowledge where beneficial. Each task can dynamically assign different weights to experts, mitigating the conflicts caused by imbalanced training.

Specifically, as shown in \Cref{fig：adjustment_model}. Given an image, let $x \in \mathbb{R}^D$ denote the shared features extracted by the ResNet backbone. Our Camera Pose Adjustment model (CPAM) consists of $M$ experts $E_m$: $\mathbb{R}^D \rightarrow \mathbb{R}^D$ and task-specific gates $G_t$: $\mathbb{R}^D \rightarrow \mathbb{R}^M$, where $t \in [1,2]$ indicates different tasks. Each gate follows the softmax design:
\begin{equation}
G_t(x)=\operatorname{Softmax}\left(\mathrm{FFN}_t(x)\right)
\end{equation}
where $\mathrm{FFN}_t$ represents a task-specific feed-forward network. The output feature $f_t$ of each task branch is computed as:
\begin{equation}
f_t=\sum_{i=1}^m G_t(x)_i \cdot E_i(x)
\end{equation}

Finally, these task-specific features $f_t$ are processed through separate MLP layers to generate the suggestion prediction $\widehat{\boldsymbol{y}}_s^i$ and the adjustment prediction $\widehat{\boldsymbol{y}}_a^i$

\textbf{Optimization.} The CPAM uses a multi-task loss function. For the suggestion prediction task, we adopt the cross-entropy loss function $\mathcal{L}_{ce}$:
\begin{equation}
\mathcal{L}_{suggest}= \frac{1}{N}\sum_{i=1}^N\mathcal{L}_{ce}(\widehat{\boldsymbol{y}}_s^i,\boldsymbol{y}_s^i)
\end{equation}
For the adjustment prediction task, the loss function is:
\begin{equation}
\mathcal{L}_{cs}=1-\frac{1}{N}\sum_{i=1}^N\frac{\widehat{\boldsymbol{y}}_a^i \cdot {\boldsymbol{y}}_a^i}{\|\widehat{\boldsymbol{y}}_a^i\|\|{\boldsymbol{y}}_a^i\|}
\end{equation}
\begin{equation}
\mathcal{L}_{norm}=\frac{1}{N}\sum_{i=1}^N(\|\widehat{\boldsymbol{y}}_a^i\|-\|{\boldsymbol{y}}_a^i\|)^2
\label{eq:L_norm}
\end{equation}
\begin{equation}
\mathcal{L}_{adjust}= \mathcal{L}_{cs} + \mathcal{L}_{norm}
\label{eq:L_adjust}
\end{equation}
The total loss function can be summarized as
\begin{equation}
\mathcal{L}_{CPAM}= \mathcal{L}_{suggest} + \mathbf{1_{(\boldsymbol{y}_s=1)}} \mathcal{L}_{adjust}
\end{equation}
where $\mathbf{1_{(\boldsymbol{y}_s=1)}}$ is an indicator function that determines during training, the gradients of the adjustment predictor are backpropagated only for samples where a suggestion should be provided.

\section{Experiments}
\label{sec:experiments}
\subsection{Implementation Details}
\textbf{Training}. We use CLIP (RN50) \cite{CLIP} as the backbone of CCQA, with RoIAlign size of $14\times14$. The CCQA model trained for 120 epochs using Adam optimizer \cite{adam} with learning rate $5 \times10^{-6}$. For CPAM, we adopt ImageNet pre-trained ResNet50 \cite{resnet} and train it for 50 epochs using Adam with learning rate $1 \times10^{-4}$ and weight decay $1\times10^{-4}$.
.

\textbf{Datasets}. We train CCQA on GAICv2 \cite{r5} (3,636 images, 86 views per image) and evaluate its generalization on CPC \cite{r4} (10,800 images, 24 views per image). Our PCARD dataset is divided into training and test sets with an 8:2 ratio for CPAM training and evaluation.

\textbf{Evaluation Metrics}. We use the AUC (Area under receiver operating characteristics curve) to evaluate the performance of the suggestion predictor. This metric measures how accurately a model triggers suggestions. Then, we evaluate the accuracy of the adjustment predictor using cosine similarity (CS) and MAE, where cosine similarity (CS) measures how close the predicted adjustment direction is to the actual adjustment direction, and MAE measures the precision of the adjustment predictor. We adopt Intersection over Union (IoU) to quantify the accuracy of view adjustment predictions. Notably, the IoU is computed on the spherical panorama surface. More details can be seen in Supplementary Material.

\subsection{Objective Evaluation}
\textbf{Exploration of different expert numbers in CPAM}. To investigate the optimal number of experts in our Camera Pose Adjustment model, we conducted ablation studies by varying the number of experts $M$ from 1 to 5. As shown in \Cref{tab:ablation_cpa}, we can observe that: $\textbf{(a)}$ the model achieves the best overall performance when $M=2$. The suggestion predictor demonstrates the highest AUC  of $79.3\%$, and the adjustment predictor shows superior performance across all metrics; $\textbf{(b)}$ increasing the number of experts beyond two leads to a gradual decline in performance across all metrics. This degradation might be attributed to the increased model complexity in expert predictions; $\textbf{(c)}$ despite having similar suggestion prediction performance (AUC scores of $78.4\%$ and $78.7\%$ respectively), the $M=3$ configuration demonstrates superior adjustment prediction capability compared to the $M=1$. This is because when $M=1$, the gating network becomes ineffective, so the CPAM lacks the dynamic expert weighting mechanism that is crucial for the mixture of experts; $\textbf{(d)}$ we also report the IoU metrics for both true positives (TP) and all predicted adjustment cases (TP+FP) from the suggestion predictor. Notably, when $M \geq2$, our adjustment predictor can still generate reasonable adjustments even when the suggestion predictor makes mistakes.

\begin{table}[t]
    \centering
    \begin{tabular}{c|c|ccc|c}
    \hline
    \multirow{2}{*}{M} & \multirow{2}{*}{AUC$\uparrow$ } & \multicolumn{3}{c|}{TP} & TP+FP \\
    \cline{3-6}  
    & & CS$\uparrow$ & MAE$\downarrow$  & IoU $\uparrow$ & IoU $\uparrow$ \\
    \hline  
    1 & \underline{78.7} & 0.401 & 0.524 & 0.604 & 0.601 \\
    2 & \textbf{79.3} & \textbf{0.415} & \textbf{0.507} & \textbf{0.613} & \textbf{0.617} \\
    3 & 78.4 & \underline{0.408} & \underline{0.515} & \underline{0.606} & \underline{0.612} \\
    4 & 77.3 & 0.398 & 0.52 & 0.597 & 0.601 \\
    5 & 76.7 & 0.368 & 0.541 & 0.591 & 0.597 \\
    \hline
    \end{tabular}
    \caption{Ablation study of Camera Pose Adjustment model. (TP: True Positive, FP: False Positive). }
    \label{tab:ablation_cpa}
\end{table}

\textbf{Exploration of different loss functions in CPAM}. To further validate the rationality of the loss functions (\cref{eq:L_adjust}) for the adjustment predictor, we compared the results of two other loss functions on the PCARD dataset, as shown in \Cref{tab:ablation_loss}. $A$ represents replacing \cref{eq:L_adjust} with MSE loss, treating the camera pose adjustment prediction as a standard regression problem. $B$ denoted replacing \cref{eq:L_norm} with MSE loss, which treats the camera pose adjustment prediction as a regression of direction and coordinates, using cosine similarity to supervise the alignment of predicted and labeled directions, and MSE for the coordinate regression. $C$ is the loss functions adopted in our paper, treating camera pose adjustment prediction as a strict spatial vector prediction, which involves supervision of both the vector direction and the vector magnitude. The results demonstrate the effectiveness of our designed loss functions. 

\begin{table}[t]
    \centering
    \begin{tabular}{c|c|ccc|c}
    \hline
    \multirow{2}{*}{Loss} & \multirow{2}{*}{AUC$\uparrow$ } & \multicolumn{3}{c|}{TP} & TP+FP \\
    \cline{3-6}  
    & & CS$\uparrow$ & MAE$\downarrow$  & IoU $\uparrow$ & IoU $\uparrow$ \\
    \hline  
    A & 76.1 & 0.391 & 0.507 & 0.602 & 0.605 \\
    B & 78.5 & 0.408 & 0.507 & 0.606 & 0.611 \\
    C & \textbf{79.3} & \textbf{0.415} & \textbf{0.507} & \textbf{0.613} & \textbf{0.617} \\
    \hline
    \end{tabular}
    \caption{Ablation Study of Loss Functions in Camera Pose Adjustment Model. (TP: True Positive, FP: False Positive)}
    \label{tab:ablation_loss}
\end{table}

\begin{figure}[t]
    \centering
    \includegraphics[width=1\linewidth]{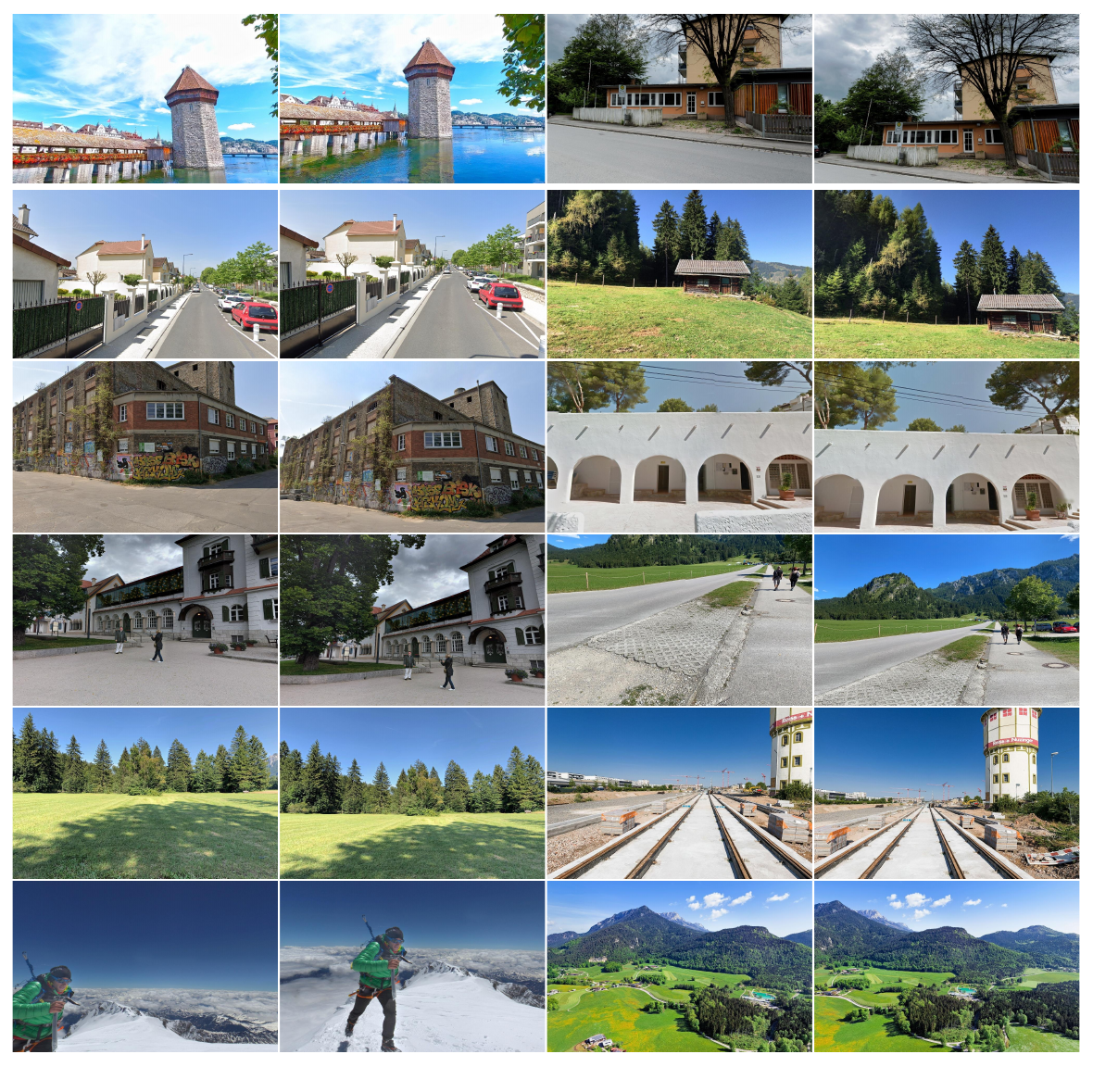}
    \caption{Qualitative examples. Each pair shows the original image (left) and the result of the adjustment (right).}
    \label{fig:qualitative_example1}
\end{figure}
\textbf{Generalization Capability Validation of CCQA}.To evaluate the generalization capability of our proposed CCQA model, and demonstrate the reliability of the scoring order in our PCARD dataset, we conducted rigorous experiments on additional unseen datasets. Specifically, we trained the model on the GAICv2 dataset\cite{r5} and then tested it directly on the unseen CPC dataset\cite{r4}. The averaged top-k accuracy ($\overline{A c c_{k}}$) and weighted average top-k accuracy($\overline{A c c_{k}^w}$) for both k=5 and k=10 as evaluation metrics are reported in \Cref{tab:comparsion_ccqa}. There are no discarded regions in the images of our PCARD dataset. Therefore, the networks were appropriately modified to adapt to our dataset, $*$ indicates that we removed RODAlign from these networks designed for image cropping tasks. The best generalization capability results are marked in bold and the second generalization capability results are marked with underlines. 

We can see that our proposed CCQA achieves the best performance on all metrics, the results demonstrate that the CCQA model exhibits good generalization capability, suggesting that utilizing this model to provide aesthetic scoring for our PCARD dataset is a reliable approach.

The ablation studies of CCQA and analysis experiments on PACRD dataset labels will be discussed in the Supplementary Material.

\begin{table}[t]
    \centering
    \begin{tabular}{@{}cccccc@{}}
        \toprule
        Method &  $\overline{A c c_5}$ & $\overline{A c c_{10}}$ & $\overline{A c c_{5}^w}$ & $\overline{A c c_{10}^w}$ \\ 
        \midrule
        TransView*\cite{r8} & \underline{51.1} & 66.4 & 36.7 & 50.7 \\ 
        GAICv2*\cite{r5} & 50.9 & \underline{66.5} & 36.6 & \underline{50.8}\\
        SFRC* \cite{r6}& 51 & 65.9 & \underline{36.8} & 50.6 \\ 
        CCQA(Ours) & \textbf{56.1} & \textbf{72.6} & \textbf{39.8} & \textbf{55.5} \\ 
        \bottomrule
    \end{tabular}
    \caption{Comparison of the generalization ability of different composition scoring models on the CPC dataset. }
    \label{tab:comparsion_ccqa}
\end{table}
\subsection{Subjective Evaluation}
\begin{table}[t]
    \centering
    \begin{tabular}{ccc}
        \toprule
        Which is better & Suggestion & Adjustment   \\ 
        \midrule
        After/Original & $82.0\%$ & $64.0\% $   \\ 
        Before/Candidate &  $14.0\%$ & $27.0\%$ \\
        No difference & $4.0\%$ &  $9.0\%$ \\
        \bottomrule
    \end{tabular}
    \caption{Subjective evaluation results on our dataset PCARD. }
    \label{tab:user_study}
\end{table}
To further demonstrate the effectiveness of our proposed framework, we design an annotation toolbox and conduct two sets of user studies. First, we select 100 images from our dataset and show the raters the image both before and after applying the suggested camera adjustment strategy to evaluate whether the suggested camera adjustment strategy effectively improves the composition of the original image. Second, we select another 50 image pairs from our dataset and show the raters the original image and candidate image to evaluate the accuracy of whether a camera pose adjustment should be suggested. To make the comparison fair, we invited 25 students to participate in the user study. The subjects are asked which image has the better composition or if they cannot tell. The order of the two images is chosen randomly to avoid bias. The results are in \Cref{tab:user_study} and the qualitative examples are shown in \Cref{fig:qualitative_example1}. When a camera pose adjustment suggestion is provided, our framework effectively improves the composition of images in most cases ($64.0\%$), with erroneous adjustment suggestions accounting for about $27.0\%$. When no suggestion is needed, our model has a high success rate ($82.0\%$), and it only wrongly judges the need for a suggestion about $14.0\%$ of the time. More qualitative results can be provided in Supplementary Material.

\section{Concluding remarks}
\label{sec:conclusion}
We have presented a new smart point and shoot (SPAS) solution to help smartphone users to take better photographs. We have made several contributions in this paper including a large dataset with $320K$ images from 4000 scenes where each image containing camera pose information. We have also developed an image quality labeler that can discern subtle image quality difference as well as a camera pose adjustment model that using a mixture of experts solution to accomplish two sequential tasks of guiding a user to compose a good shot of a scene.

\section*{Acknowledgment}

This work was supported in part by the National Natural Science Foundation of China under Grant 62271323 and U22B2035, in part by Guangdong Basic and Applied Basic Research Foundation under Grant 2023A1515012956 and 2023B1212060076, and in part by the Shenzhen Research and Development Program under Grant JCYJ20220531102408020 and KJZD20230923114209019.
{
    \small
    \bibliographystyle{ieeenat_fullname}
    \bibliography{main}
}


\end{document}